\documentclass[12pt]{article}

\usepackage{hyperref}
\usepackage{url}
\usepackage[latin9]{inputenc}
\usepackage{float}
\usepackage{mathtools}
\usepackage{bm}
\usepackage{amsmath}
\usepackage{amsthm}
\usepackage{amssymb}
\usepackage{amsfonts} 
\usepackage{comment}
\usepackage{graphicx}
\usepackage{subcaption}
\usepackage{psfrag,epsf}
\usepackage{enumerate}
\usepackage{natbib}

\providecommand{\tabularnewline}{\\}
\floatstyle{ruled}
\newfloat{algorithm}{tbp}{loa}
\providecommand{\algorithmname}{Algorithm}
\floatname{algorithm}{\protect\algorithmname}

\theoremstyle{plain}
\newtheorem{thm}{\protect\theoremname}
\theoremstyle{remark}
\newtheorem{rem}[thm]{\protect\remarkname}
\theoremstyle{plain}

\usepackage{psfrag}\usepackage{epsf}\usepackage{enumerate}
\usepackage{natbib}
\usepackage{amsfonts}\usepackage{bm}
\usepackage{color}
\usepackage{colortbl}
\usepackage{algorithm}
\usepackage{algorithmic}
\usepackage{balance}
\usepackage{multirow}
\usepackage{cuted}

\def\th@plain{%
  \thm@notefont{}
  \itshape 
}
\def\th@definition{%
  \thm@notefont{}
  \normalfont 
}

\DeclareMathOperator*{\argmin}{arg\,min}

\newcommand{\rom}[1]{\uppercase\expandafter{\romannumeral #1\relax}}
\DeclarePairedDelimiterX{\norm}[1]{\lVert}{\rVert}{#1}
\DeclarePairedDelimiterX{\bnorm}[1]{\biggl\lVert}{\biggr\rVert}{#1}
\DeclarePairedDelimiterX{\abs}[1]{\lvert}{\rvert}{#1}

\renewcommand{\arraystretch}{1.5} 
\renewcommand{\baselinestretch}{1} 

\newtheorem{lemma}{Lemma}\newtheorem{assumption}{Assumption}

\def\T{{ \mathrm{\scriptscriptstyle T} }}
\def\P{{ \mathrm{pr} }}
\def\v{{\varepsilon}}

\def\F{\textsc{F}}
\def\E{\mathbb{E}}

\def\F{\mathcal{F}}
\def\P{\mathbb{P}}
\def\D{\mathcal{D}} 
\def\Mp{\mathcal{M}^{p}} 
\def\Mn{\mathcal{M}^{non}} 
\def\M{\mathcal{M}} 
\def\Ej{\mathbb{E}_{n_j}}

\def\lap{\mathtt{L}_{\mathtt{S}}}
\def\S{\mathcal{S}}

\newcommand{\REV}[1]{{#1}}  

\providecommand{\theoremname}{Theorem}

\providecommand{\remarkname}{Remark}
\providecommand{\theoremname}{Theorem}

\providecommand{\corollaryname}{Corollary}
\providecommand{\remarkname}{Remark}
\providecommand{\theoremname}{Theorem}

\global\long\def\var{\textrm{Var}}%

\title{Meta Clustering for Collaborative Learning }

\newcommand{\blind}{0}

\begin{document}
\def\spacingset#1{\renewcommand{\baselinestretch}%
{#1}\small\normalsize} \spacingset{1}

\if0\blind
{
  \title{\bf Meta Clustering for Collaborative Learning}
  \author{Chenglong Ye\\
    Dr. Bing Zhang Department of Statistics, University of Kentucky\\
    Reza Ghanadan\hspace{.2cm}\\
    Google Research\\
    and \\
    Jie Ding
    \\
    School of Statistics, University of Minnesota\\
    }
  \maketitle
} \fi

\if1\blind
{
  \bigskip
  \bigskip
  \bigskip
  \begin{center}
    {\LARGE\bf Title}
\end{center}
  \medskip
} \fi

\bigskip

\begin{abstract}
In collaborative learning, learners coordinate to enhance each of their learning performances. From the perspective of any learner, a critical challenge is to filter out unqualified collaborators. We propose a framework named meta clustering to address the challenge. Unlike the classical problem of clustering data points, meta clustering categorizes learners. Assuming each learner performs a supervised regression on a standalone local dataset, we propose a Select-Exchange-Cluster (SEC) method to classify the learners by their underlying supervised functions. We theoretically show that the SEC can cluster learners into accurate collaboration sets. Empirical studies corroborate the theoretical analysis and demonstrate that SEC can be computationally efficient, robust against learner heterogeneity, and effective in enhancing single-learner performance. Also, we show how the proposed approach may be used to enhance data fairness. Supplementary materials for this article are available online.

\end{abstract}
\noindent%
{\it Keywords:}  Distributed computing; Data Integration; Fairness; Meta clustering; Regression.
\vfill

\newpage
\spacingset{1.5} 
\renewcommand{\arraystretch}{1}

%

\newcommand{\fix}{\marginpar{FIX}}
\newcommand{\new}{\marginpar{NEW}}

\section{Introduction}

Collaborative learning has been an increasingly important area that
aims to build a higher-level, simpler, and more accurate global model
by combining various sources. 
The data from each source can be regarded as a sub-dataset of
an overarching dataset. These sub-datasets are usually heterogeneous
and stored in decentralized locations for various reasons. For example,
each sub-dataset is from a unique research activity with domain-specific
features, data are too large to be stored in one location, or the data privacy concern entails separate access to sub-datasets. Suppose each sub-dataset is handled by a {\it learner}. A natural way to improve the modeling performance is to integrate these learners to leverage the distributed
computing resources and enlarged sample size. 

\begin{figure}[tb]
\centering \includegraphics[width=1\linewidth]{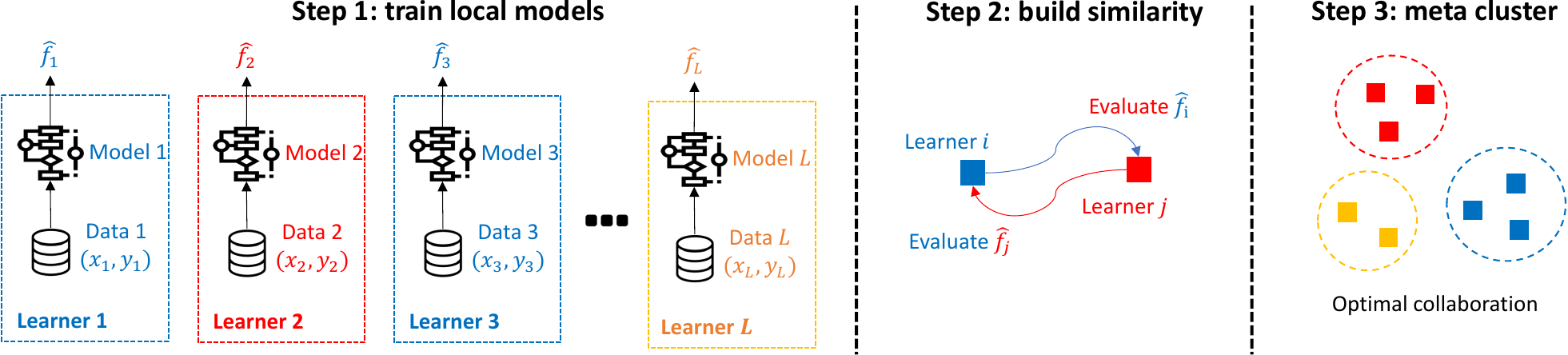}
\vspace{-0.3in}
 \caption{{\small Illustration of meta clustering for learners/datasets,
based on supervised relationships.}}
\label{fig_overview} \vspace{-0in}
 
\end{figure}

The general question of ``how to collaborate'' has led to several recent research on collaborative learning, which we will elaborate in Section~\ref{subsec:Related-work}. This paper aims to answer the following question: Whom to collaborate with? Selecting collaborators is crucial when not all learners are qualified, such as learners with incapable models or irrelevant sub-datasets. In particular, we
suppose each sub-dataset is of a supervised nature, consisting of predictor-response pairs $(  \bm{X}, Y)$. A learner tends to collaborate
with those whose data exhibit the same or similar underlying $  \bm{X}$-$Y$ relationship. To that end, we propose to study the
problem of \textit{clustering for supervised relationships}. The idea is that
sub-datasets exhibiting similar function relationships (between
$  \bm{X}$ and $Y$) should fall into the same cluster. 
An alternative view
of such clustering is categorizing sub-datasets into fewer meta-datasets,
offering better learning quality without inducing many estimation biases. As such, we name the problem ``meta clustering.'' Unlike the classical learning problem of data-level clustering, our goal here is to cluster
datasets instead of single data points. In this framework, learners should collaborate with those in the same cluster. 
We focus on the regression scenario, where each sub-dataset can
be modeled by $f(  \bm{X})=E(Y|  \bm{X})$ for some function $f$, and sub-datasets in the same cluster share the same (latent) function $f$. 
We propose a computationally efficient algorithm for meta clustering, consisting of three steps: select, exchange, and cluster. Figure~\ref{fig_overview} illustrates the main idea of the proposed method. In summary, we first train local models for each learner and select the best model. Then, each pair of learners exchange their already-learned best models. We then calculate the similarity between each pair of two learners by evaluating one's model on the other's dataset. Finally, spectral clustering is performed based on the similarity matrix. 

The contribution of our work is three-fold. First, we propose to study the problem of clustering for datasets based on the underlying supervision relationships. The problem of meta clustering naturally fits the emerging need for robust collaborations in adversarial learning scenarios. We propose a general approach named Select-Exchange-Cluster (SEC).
Second, the proposed SEC method is both computationally efficient and theoretically guaranteed. We show that when the sample size of each sub-dataset is sufficiently large, the sub-datasets with the same generating function can be accurately categorized into the same cluster. Moreover, the number
of clusters does not need to be specified in advance, and it can be appropriately identified in a data-driven manner. Third, we can use the proposed method
in general supervised regression tasks that involve non-linear
and nonparametric learning models. It can be used for various learning tasks even if learners are not sure about the existence
of latent functions. For example, we show its use to significantly enhance
the prediction performance under data fairness constraints, where a reduction of approximated 50\% prediction error is achieved without using any sensitive variable.

\subsection{Related work\label{subsec:Related-work}}

We briefly describe the connection between meta clustering and existing research.

\textit{Collaborative learning.} When data are stored across distributed
clients such as edge devices, directly sharing local datasets compromises data
privacy. Federated learning~\citep{konevcny2016federated,mcmahan2016communication,FLoutlook} is a popular collaborative learning framework that aims to train a global model on distributed datasets without sharing local data. The main idea is to exchange model parameters updated from local data and iteratively update the globally trained model (assuming the same model for all clients). More general federated learning frameworks beyond exchanging parameters have been recently developed~\citep{HeteroFL,SemiFL}. Our proposed meta clustering framework may
serve as a preliminary analysis tool for selecting ``qualified'' collaborators before applying any federated learning algorithm. 
Assisted learning~\citep{Assist,GAL,Recommender} is another recently developed collaborative learning framework for decentralized organizations, where any organization being assisted or assisting others does not share its local data, model, or learning objective. In assisted learning, data variables held by participants are often distinct and assumed to be linked by a non-private identifier. In contrast, our paper focuses on the scenario where participants have the same variables, but the supervised relationships are possibly heterogeneous.

\textit{Data Integration}. Data integration aims to improve statistical
performance by sharing model parameters or combining datasets. Many methods have
been proposed in this research direction. For example, \citet{tang2016fused} developed a fused
lasso approach to learn parameter heterogeneity in linear models
on different datasets. \citet{li2018integrative} proposed an integrative
method of linear discriminant analysis (LDA) for multi-type data,
which was shown to improve classification accuracy over
the performance on a single dataset. \citet{jensen2007bayesian}
proposed a Bayesian hierarchical model in a variable selection framework
that integrates three types of data in gene regulatory networks: gene
expression, ChIP binding, and promoter sequence.
\REV{\citet{yang2019high} studied the problem of integrating regression data from different sources by pooling data for centralized learning. They proposed an objective function that estimates regression coefficients by penalizing pairwise differences between coefficients of the same covariate to identify heterogeneous and homogeneous coefficients automatically.
\citet{hector2020joint} proposed a method for joint integrative analysis of multiple data sources with correlated vector outcomes under a distributed quadratic inference function framework. They assume the clustering of data sources is known. In that regard, our approach may be used as a preliminary step before applying their method when the underlying clustering structure is unknown.}

In comparison with most data integration methods where statistical models
are specified in each sub-dataset, our proposed meta clustering framework is model-free in the sense that it allows each learner to use different
local models without sharing the form of those models. For example,
one learner can use a linear model to fit a sub-dataset, while
another can use a random forest. The proposed SEC algorithm only exchanges the predicted
values for clustering without exchanging the parameters or the models. 
\REV{It is worth noting that with our meta clustering, a learner considers a binary decision whether to collaborate with
another learner or not. A similar setup was also considered by \citet{CollabParam}, where the authors proposed the notion of model linkage selection for learners who share parameters of common interest. Alternatively, a learner may use a soft decision-based collaboration with others. In that direction, \citet{shen2020fusion} developed an approach that summarizes inference results from other learners as confidence density functions and then combines them using a weighting scheme. \citet{tan2021tree} proposed a tree-based ensemble approach that integrates the prediction results from other learners as feature variables.
}

\textit{Divide-and-conquer.} Divide-and-conquer in the context of distributed learning often refers to the procedure that partitions a large
dataset into sub-datasets and then combines results (e.g., \textit{p}-values,
coefficients) obtained from each sub-dataset. 
For example, \citet{JMLR:v16:zhang15d} proposed a method
that randomly partitions the dataset into sub-datasets and fits a
kernel ridge regression estimator in each sub-dataset. A simple average of local predictors is used as the global estimator, achieving minimax optimal convergence rates. \citet{JMLR:v16:mackey15a}
proposed the {D}ivide-{F}actor-{C}ombine (DFC)
framework for noisy matrix factorization, which improves the scalability
and enjoys estimation guarantees. \citet{fan2019distributed} proposed
a distributed {P}rinciple {C}omponent {A}nalysis (PCA)
algorithm for data stored across multiple locations, which performs
similarly to the PCA estimator based on the whole dataset. Different
assumptions of the distributed sub-datasets were also investigated,
such as independent cross-sectional data \citep{doi:10.1198/jasa.2011.tm09803},
independent sources/studies \citep{doi:10.1080/01621459.2014.957288,battey2015distributed},
network meta-analysis \citep{YANG2014105}, high-dimensional correlated
data \citep{hector2020distributed}, and multi-measurements data from different experiments \citep{gao2017data}.

The primary goal of divide-and-conquer is to reduce computational costs via
parallel computing across sub-datasets. One learner may or may not have access to all the sub-datasets. In our framework, each learner can only
access its local sub-dataset. Also, divide-and-conquer
methods assume the underlying relationship between the response
and the predictors for each sub-dataset is the same, so combining results from all the sub-datasets is reasonable. 
However, the datasets in distributed storage may be heterogeneous in distributions.  
Identifying the potential clustering of the subs-datasets is important for bias reduction and robust modeling. For divide-and-conquer methods, meta clustering can be applied to \REV{analyze whether there
are potential cluster structures} on the whole dataset. If there exist cluster structures, a random splitting in divide-and-conquer
may lead to a modeling bias.

The remainder of the paper is outlined below. We describe the meta clustering problem in Section \ref{sec:ProblemSetup} and propose our method, together with its theoretical properties, in Section \ref{sec:Method}. In Section \ref{sec:Data-Fairness},
we demonstrated a potential use of the method in fairness learning
scenarios. In Sections \ref{sec:Simulations} and \ref{sec:Real-Data-Applications},
we show the performance of our method through more experimental studies.
The proofs are included in the supplementary material.

\section{Problem\label{sec:ProblemSetup}}

Suppose the dataset $\D:=\{\D_{i}\}_{i=1}^{L}$ is the union of $L$
sub-datasets. For example, $\D_{i}$ can represent
the sub-dataset stored in the  $i$-th location/server, the sub-dataset
from the  $i$-th study in a meta-analysis, or the sub-dataset from
the  $i$-th patient in the same research project. We assume each sub-dataset
$\D_{i}$ is handled by a learner $l_{i}$ who considers a set of
available methods $\M_{i}=\Mp_{i}\cup\Mn_{i}$ for data analysis.
Here $\Mp_{i}$ ($\Mn_{i}$) denotes the parametric (nonparametric)
models in $\M_{i}$.
Briefly speaking, we assume a parametric model (e.g., a linear regression model) has a better convergence rate than a nonparametric one (e.g., a decision tree), and the latter is consistent in estimation. More detailed assumptions are included in the supplementary document.
The notions of parametric and nonparametric are made only for technical convenience. It is practically hard to distinguish them with finite samples, even in linear models. We refer to~\citep{DingOverview} for more discussions on this. 
Parallel computing can be regarded as a particular case when all sub-datasets use the same learner.
Throughout the paper, \REV{we will use lowercase letters (e.g., $x$, $\bm{x}$, $a_n$) to denote observed data or constants, uppercase letters to denote random variables or vectors (e.g., $X$, $\bm{X}$), typewriter uppercase letters (e.g., $\mathtt{A}$) to represent matrices, and calligraphy uppercase letters (e.g., $\mathcal{A}$) to represent sets.}

Suppose the sub-dataset $\D_{i}$ consists of $n_{i}$ independent
data points, denoted by $\D_{i}=\{(y_{i,j},\bm{x}_{i,j}):y_{i,j}\in\mathbb{R},\bm{x}_{i,j}\in\mathbb{R}^{p}\}_{j=1}^{n_{i}}$,
from the underlying model 
$ 
Y_{i}=f_{i}(  \bm{X}_{i})+\v_{i}, 
$ 
where $  \bm{X}_{1},\ldots,  \bm{X}_{L}$ are independent $p$-dimensional random variables with a distribution function $P_{\bm{X}}(\cdot)$,
and the noise $\v_{i}\sim \mathcal{N}(0,\sigma_{i}^{2})$ is independent of
$  \bm{X}_{i}$. Moreover, for any $i_{1},i_{2}\in\{1,\ldots,L\}$, $\v_{i_{1}}$
is independent of $  \bm{X}_{i_{2}}$. We suppose the $L$
sub-datasets consist of the same $p$ predictors. 

Let $n:=n_{1}+\cdots+n_{L}$ denote the overall sample size. Throughout the paper, we assume
there are $K$ (fixed but unknown) data generating functions, namely
$f_{i}\in\F=\{f^{(1)},\ldots,f^{(K)}\}$ for $i=1,\ldots,L$. Let $||\cdot||$ denote the Euclidean norm. 
Define the $L_{2}$ norm $\norm{f}_{2}=\sqrt{\int f(\bm{x})^{2}P_{  \bm{X}}(d  \bm{x})}$
and the $L_{\infty}$ norm $||f||_{\infty}=\textrm{ess}\sup|f|=\inf\{c\geq0:|f(  \bm{X})|\leq c\ \textrm{a.s.}\}$.
We say two underlying models $f^{(i)}$ and $f^{(j)}$ are different if $||f^{(i)} - f^{(j)}||_{\infty}>0$. 

Our goal is to  accurately cluster the $L$ sub-datasets into $K$ clusters,
where the underlying regression functions corresponding to the sub-datasets in the same cluster are similar.

\section{Method\label{sec:Method}}

The intuition of our method is that if two sub-datasets are from the same or similar data generating function, a modeling procedure should produce similar results on the two sub-datasets. We propose the following three-step method named  \textbf{S}elect-\textbf{E}xchange-\textbf{C}luster
(SEC), where learners communicate with their estimated regression functions.

\noindent \textbf{Step 1 [Select]}:
Each learner uses
its own sub-dataset to learn a model from a set of candidate methods $\M_{i}$. 
Suppose each learner conducts the half-half cross-validation to perform model selection. 
In particular,  learner $l_{i}$
splits the data $D_{i}$ into two parts $\D_{i,1}$ and $\D_{i,2}$
of equal size $n_{i,1}=n_{i,2}=n_{i}/2$ (assuming an even $n_{i}$ for simplicity). The learner applies each candidate method
$\delta\in\M_{i}$ to the training set $\D_{i,1}$ and obtains the
corresponding estimator $\hat{\delta}_{n_{i,1}}$. For learner $l_{i}$,
denote the best method $\delta_{i}$ as the one that minimizes the
mean squared error (MSE) on the test set $\D_{i,2}$ , namely 
\begin{equation}
\delta_{i}=\argmin_{\delta\in\M_{i}}\sum_{(y,\bm{x})\in\D_{i,2}}\bigl(y-\hat{\delta}_{n_{i,1}}(\bm{x})\bigr)^{2}/n_{i,2}.\label{eq_CV}
\end{equation}
The ``best'' method $\delta_{i}$ is then applied to the whole data
$\D_{i}$ to estimate the underlying function $f_{i}$. Denote the
resulting estimated function as $\hat{f}_{{i}}$ and its fitted
mean squared error as 
$
\hat{e}_{i}:=\sum_{(y,\bm{x})\in\D_{i}}\bigl(y-\hat{f}_{{i}}(\bm{x})\bigr)^{2}/n_{i}.
$ To summarize, for each learner $l_{i}$, we have the non-shared data $\D_{i}$ and the sharable information
$\{\delta_{i},\hat{f}_{{i}},\hat{e}_{i}\}$. 

\noindent \textbf{Step 2 [Exchange]}:
For any two learners, they exchange the sharable
information $\{\delta_{i},\hat{f}_{{i}},\hat{e}_{i}\}$. 
In particular, denote
$v_{ij}$ as the dissimilarity between any two learners $(l_{i},l_{j})$, $i\neq j$. We apply the  $i$-th learner's best estimator $\hat{f}_{{i}}$
to the  $j$-th learner's dataset $\D_{j}$ and obtain its prediction
loss
$\hat{e}_{i\rightarrow j}:=n_{j}^{-1}\sum_{(y,\bm{x})\in\D_{j}}\bigl(y-\hat{f}_{{i}}(\bm{x})\bigr)^{2},
$
where the subscript $i\rightarrow j$ denotes the information flow from $l_{i}$ to $l_{j}$. Similarly, we apply $\hat{f}_{{j}}$
to the dataset $\D_{i}$ and obtain the prediction loss $\hat{e}_{j\rightarrow i}$.
The dissimilarity $v_{ij}$ is then defined as the difference between
their best estimators: 
\begin{equation}
v_{ij}=|\hat{e}_{i\rightarrow j}-\hat{e}_{j}|+|\hat{e}_{j\rightarrow i}-\hat{e}_{i}|,\label{eq:v}
\end{equation}
where  $v_{ij}=v_{ji}$ for any $i\neq j$. When $i=j$, the
self-dissimilarity of a learner $l_{i}$ is $v_{ii}:=0$. 

\noindent \textbf{Step 3 [Cluster]}:
Based on the dissimilarity $v_{ij}$, a similarity
matrix is constructed, which is used to cluster the $L$ learners. 
In particular, we calculate a symmetric matrix $\mathtt{S}$ whose $(i,j)$-th component is
$ \mathtt{S}_{ij}:=\exp(-av_{ij})$. 
Here, $a$ is a tuning parameter for computational
convenience. For example, when $\min_{i,j}v_{ij}$ is large and $a=1$, $ \mathtt{S}_{ij}$'s can be
negligibly small for all $(i,j)$ and thus become not distinguishable by the computer (due to its limited precision).  
Let $\mathcal{P}=\{1,\ldots, L\}$ denote the set of labels of the $L$ learners.
For a given $K$, we will find a collection of sets \REV{$\{\S_{i}\}_{i=1}^{K}$ that forms a partition of $\mathcal{P}$}. The partition is obtained by applying a spectral clustering algorithm to the matrix $ \mathtt{S}$ and dividing the $L$ learners into $K$ groups.
For completeness, we summarize the clustering step (Step 3) in Algorithm~\ref{algo_main}. 
 
\begin{algorithm}[tb]
\footnotesize
\textbf{Input}: Number of learners $L$, learners/datasets \{$\D_{i}\}_{i=1}^{L}$,
the number of clusters $K$ (optional).

\textbf{Output}: The number of clusters $K$ (if not given), and the cluster labels $c_{i}\in \{1,\ldots,K\},i=1,\ldots,L$. 
\begin{enumerate}
\item Calculate the similarity matrix $ \mathtt{S}\in\mathbb{R}_{+}^{L\times L}$,
where each $ \mathtt{S}_{ij}=\exp(-av_{ij})$ and $v_{ij}$ is given by \eqref{eq:v}. 
\item If $K$ is given, conduct the spectral clustering: 
\begin{enumerate}
\item Calculate the \REV{Laplacian $\lap$ of $ \mathtt{S}$}: %
$ \lap= \mathtt{D}^{-1/2} \mathtt{S} \mathtt{D}^{-1/2}$, where $\mathtt{D}:=\text{diag}(\sum_{j=1}^{L} \mathtt{S}_{1j},\ldots, \sum_{j=1}^{L} \mathtt{S}_{Lj})$. 
\item Compute the $K$ largest eigenvectors of $\lap$: $\bm{u}_{1},\ldots,\bm{u}_{K}$.
Denote $ \mathtt{U}=[\bm{u}_{1},\ldots,\bm{u}_{K}]\in\mathbb{R}^{L\times K}$. 
\item Standardize each row of $ \mathtt{U}$ to have unit $\ell_{2}$ norm. Denote
the standardized matrix as $ \mathtt{U}_{*}$. 
\item Apply $\textit{k}$-means clustering to the rows of $ \mathtt{U}_{*}$ into $K$
clusters, and record the labels $c_{i},i=1,\ldots,L$. 
\end{enumerate}
\item If $K$ is not given: 
\begin{enumerate}
\item Sort the eigenvalues of $ \mathtt{S}$ from small to large and determine $K$ (Remark~\ref{selectK}). 
\item Go back to Step~2. 
\end{enumerate}
\end{enumerate}
\caption{{\small Pseudocode for the Step 3 of SEC algorithm \label{algo_main}}}
\end{algorithm}
\begin{rem}[Spectral clustering]
 There are different variants of spectral clustering in
the literature. Due to technical convenience, we build on the work of \citep{ng2002spectral}. 
We will show that the spectral clustering algorithm based on the constructed similarity matrix can guarantee desirable performance. 
\end{rem}
\begin{rem}[Selection of $K$]\label{selectK}
When $K$ is unknown, we may add a penalty term $K\cdot\lambda_{n}$
in to the $k$-means clustering in Step 2(d) of Algorithm~\ref{algo_main} to minimize 
\begin{equation}
\sum_{t=1}^{K}\sum_{i,j\in \S_{t}}\frac{1}{2|\S_{t}|}|| \mathbf{u}_{(i)}- \mathbf{u}_{(j)}||^{2}+K\cdot\lambda_{n}\label{eq:penalty}
\end{equation}
over all possible partitions of $\mathcal{P}$ and a grid of values of
$K$.
Here, $ \mathbf{u}_{(i)}$ denotes the $i$-th row of $ \mathtt{U}_*$ defined in Algorithm~\ref{algo_main}. 
The minimization problem \eqref{eq:penalty} is equivalent to comparing the within-cluster distance over a grid of $ K $ values. We suggest $\lambda_n=O(\max(n^{-1},u^4_{n}))$, where $u_n$ is an upper bound of the convergence rates of non-parametric estimators (elaborated in the supplementary document).
In practice, picking
an appropriate penalty term may be complex because of the known convergence rates of nonparametric methods in Step~1.
An alternative approach we suggest is using the gap statistics~\citep{tibshirani2001estimating} that searches for the so-called ``elbow point'' in the curve of the sum of within-cluster mean-squared errors (namely the first term in (\ref{eq:penalty})) against different $K$'s. We will also show in the supplementary document that an adequately chosen penalty can select the correct $K$ with a high probability.
 \end{rem}
\REV{
\begin{rem}[Future prediction]
The clustering
results may also be used for downstream collaborative
learning methods, where a learner only interacts with others in
the same cluster. Though prediction is not the main focus of this
paper, we discuss two use cases to perform prediction based on
the clustering results from SEC. For any particular learner $l_i$,
suppose it belongs to the cluster $\mathcal{S}_{t}$. In the first case,
the sub-datasets cannot be pooled due to communication bandwidths or privacy regulations. To collaborate, other learners in the same cluster may transmit their learned models $\hat{f}_{n_{j}}$ ($j\in \mathcal{S}_{t}, j\neq i$) to the
learner $l_i$. Then, to predict for a future observation
$\bm{x}$, the learner $l_i$ uses the weighted average of the fitted models from learners
in the same cluster, e.g.,
\begin{equation}
\sum_{i\in\mathcal{S}_{t}}\frac{n_{i}}{\sum_{j\in\mathcal{S}_{t}}n_{j}}\hat{f}_{{i}}(\bm{x}),\label{eq:futurepred}
\end{equation}
where the weights are proportional to the sample size. In this way, the above case does not
require direct data-sharing among learners. It is worth mentioning that the weights
in $\eqref{eq:futurepred}$ may not be optimal for a statistical
gain of prediction accuracy. We include further discussions on the statistical
gain in the supplementary material. 
The second use case is when the sub-datasets are allowed
to be pooled. Then, the learner $l_i$ pools all the sub-datasets in a cluster
$\mathcal{S}_{t}$ and fits one model to make future
predictions. In this case, the learner $i$ directly obtains a larger sample
and thus tends to learn a better model. Nevertheless, this case requires
the learners to share data, which may violate the purpose of collaborative
learning.
\end{rem}

}

The following theorem shows that the SEC  can accurately identify the correct clusters when the overall sample size goes to infinity. Its proof is included in the supplementary document.

\begin{thm}\label{thm}
	Under some assumptions (elaborated in the supplementary document), the labels $c_{1},\ldots,c_L$ produced by SEC satisfy  $c_i=c_j$ if and only if  $f_{i}=f_{j}$, for any $i,j$, with probability going to one as $n\rightarrow\infty$. 
\end{thm}
 \begin{rem}[Data independence]
 The clustering accuracy in the theorem may no longer hold if the independence of $y_{ij}$'s breaks down. For longitudinal settings, for example, we may assume additional conditions on $y_{ij}$ (e.g., a $\alpha$-mixing sequence) for the proof to hold. We leave the more sophisticated analysis for dependent data as future work.
 \end{rem}

\section{Application to Data Fairness\label{sec:Data-Fairness}}

One promising application of the proposed method is to enhance data fairness. Biases inherent in data collection
and techniques based on these data will not
address (sometimes even worsen) the inequity for disadvantaged
groups. In recent years, there have been many works to define fairness,
discover unfairness, and apply algorithms to promote fairness.
For example, based on the maximum likelihood principle,
\citet{kamishima2011fairness} proposed a prejudice remover regularizer
(based on the mutual information between response and
sensitive variables) for classification models.
\citet{NIPS2016_6374}
proposed a criterion called equal opportunity (or equalized odds) for a particular sensitive variable and demonstrated how
to adjust a predictor to alleviate discrimination. 
\citet{zafar2017fairness} devised a notion called positive rate disparity and proposed a method to reduce disparities in mistreatment and treatment. \citet{10.1145/3194770.3194776} compared the differences among 20 fairness definitions for classification problems. 

We consider a linear regression setting where the sensitive variable
is independent of other variables. In particular, we generate a
dataset $\D$ that consists of 50 sub-datasets $\{\D_{i}\}_{i=1}^{50}$,
each with size $n_{i}=50$ from the linear model: $Y_i=X_{1i}+2X_{2i}-2X_{3i}+2X_{4i}+b R_{i}+\epsilon_i$,
where $(X_{1i},X_{2i},X_{3i},X_{4i})\sim \mathcal{N}(0,\mathtt{I}_{4})$ are the non-sensitive variables,
$\epsilon_i\sim \mathcal{N}(0,1)$ is the random noise, and $R_{i}$ is the sensitive
variable that may induce unfairness if it were known. We consider different scales of the coefficient of
the sensitive variable, $b\in\{0.01, 0.5,1,2,3,4,5,6,20\}$. The sensitive
variable $R_{i}$ is generated from a standard normal $\mathcal{N}(0,1)$ distribution
and is set to be fixed for each given $i$. Using a fixed
value as a sensitive variable is reasonable 
for data fairness problems where multiple measurements exist for
the same subject. For example,
if $\D$ represents longitudinal data and each sub-dataset represents a
person, then the subject-specific sensitive variable (e.g., gender, race,
age, home location) is fixed for each person. We set $R_{i}$ as a continuous variable in this example. We split the dataset into a training set of 30 sub-datasets
(e.g., $\{\D_{i}\}_{i=1}^{30}$) and a test set of 20 sub-datasets (e.g.,
$\{\D_{i}\}_{i=31}^{50}$). For each learner in the test set, we further
split it into two parts of the same size (e.g., $\D_{i}=\D_{i}^{1}\cup\D_{i}^{2}$).
The splitting is because we need extra data points
to cluster the learners in the test set. Then, the dataset $\D$ is
reorganized into the following three sets: the training set $\{\D_{i}\}_{i=1}^{30}$,
the test set $\{\D_{i}^{1}\}_{i=31}^{50}$, and the validation set
$\{\D_{i}^{2}\}_{i=31}^{50}$.The random data splitting is repeated
100 times.

For the training set, in the existence of a sensitive variable, we
consider three methods of building a model: \textit{Oracle}, \textit{Fairness}, \textit{SEC-Fairness}.
The Oracle method directly builds a linear regression model from
the training set using the sensitive variable (namely without considering
fairness constraints), which is expected to have the best predictive performance. The
Fairness method builds a linear regression model on the training
set without using the sensitive variable since using the sensitive variable
is not allowed or even available in the modeling procedure. The SEC-Fairness method finds potential
groupings among the sub-datasets in the training set before building
models without the sensitive variable. In particular, it first uses
the SEC algorithm on the training set to cluster these $30$ learners
$\{\D_{i}\}_{i=1}^{30}$ into groups. \REV{Then, it uses the similarity
between a learner and a cluster to identify which cluster (identified from the training set) each of these $20$ learners in the test set belongs to. To measure the similarity
between a learner $l_{i}$ and a cluster, we use the sum of the similarities
between $l_{i}$ with each learner in that cluster. Then, the learner
$l_{i}$ belongs to a cluster if its similarity to the cluster is
larger than any other cluster.} In the SEC algorithm, for simplicity, each learner $l_{i}$ considers
two candidate modeling methods: Random Forest \citep{Breiman:2001wf}
(RF) and linear regression (LR), namely $\M_{i}=\{\textrm{RF},\textrm{LR}\}$ in the ``select'' step.
\begin{table}[tb]
\small
\begin{centering}
\begin{tabular}{ccccc}
\hline 
 & \multicolumn{2}{c}{SEC-Fairness} & \multicolumn{1}{c}{Fairness} & Oracle\tabularnewline
\hline 
$b$ & MSE  & \multicolumn{1}{c}{$\hat{K}$} & \multicolumn{1}{c}{MSE} & MSE\tabularnewline
\hline 
0.01 & 1.03 (0.005) & 1 (0) & 1.03 (0.005) & 1.03 (0.005)\tabularnewline
0.5 & 1.12 (0.007) & 2.25 (0.59) & 1.20 (0.007) & 1.03 (0.005)\tabularnewline
1 & 1.41 (0.02) & 2.65 (0.59) & 2.04 (0.03) & 0.92 (0.005)\tabularnewline
2 & 1.96 (0.04)  & 2.64 (0.48) & 5.19 (0.09)  & 1.06 (0.005)\tabularnewline
3 & 4.31 (0.22) & 2.77 (0.44) & 12.55 (0.27) & 0.96 (0.005)\tabularnewline
4 & 4.79 (0.28) & 2.86 (0.37) & 15.16 (0.36) & 1.00 (0.005)\tabularnewline
5 & 4.21 (0.12) & 2.97 (0.17) & 18.89 (0.36) & 0.99 (0.005)\tabularnewline
6 & 12.03 (0.60) & 2.81 (0.39) & 50.58 (1.04) & 1.06 (0.004)\tabularnewline
20 & 80.47  (4.90) & 2.91 (0.29) & 438.64 (9.38) & 1.07 (0.006)\tabularnewline
\hline 
\end{tabular}
\par\end{centering}
\caption{\REV{{\small Predictive performances of the three methods for the data fairness
example. (The values in the parentheses are the standard error of the
averaged MSE and the standard deviation of the estimated number of
clusters $\hat{K}$ respectively over $100$ replications.)\label{tab:fairness}} }}
\end{table}
For the validation set, we evaluate the predictive performances of
the models by the mean square error (MSE), which are presented in
Table \ref{tab:fairness}. As shown in the table, when the importance
(coefficient $b$) of the sensitive variable is high, the SEC-Fairness
 reduces the MSE of Fairness by about 50\% overall. One possible reason
is as follows. The linear relationship between $y$ and the variables
$\{X_{1},\ldots, X_{4}\}$ only differs in the intercept per
learner. The similarity between two learners, as in the SEC algorithm,
will be small if the difference between their sensitive variables $|R_{i}-R_{j}|$
is large. It is then more likely that SEC divides those with similar
values of the sensitive variable into the same cluster. It is worth mentioning that the SEC-fairness method satisfies the fairness constraint since
it does not utilize the sensitive variable at all, and the non-sensitive variables used for clustering are independent of the sensitive variable.

The Oracle method, as expected, is very stable in MSE (around 1) over different
values of $b$. When $b$ is large, SEC-Fairness is
comparable to the oracle method, though
it performs better than the Fairness method. One reason is that the
estimated number of clusters $\hat{K}$ is in the interval
$[2,3]$. In this example, we select $\hat{K}$ by the gap statistic. We note that there exists no ``true'' value of $K$ since every learner/sub-dataset
has a unique sensitive value. 
\REV{In the case $b=20$, we have $\hat{K}=2.91$. But if we force $
\hat{K}=10$
in the SEC algorithm, the MSE performance of SEC-Fairness
is much improved. One reason that the gap statistic selects a small $
\hat{K}$ is that it chooses the
value of $K$ that most reduces the gap compared with $K-1$
instead of selecting $K$ that achieves the global minimum.
Consequently, the gap statistic tends to select $\hat{K}$ as 3 or 4 in this
example. An alternative way to estimate $K$ is
cross-validation. Specifically, we can split each sub-dataset into
a training set and a test set. Then, on the collection of the test
sets, we can compare the MSE performance based on a list of $K$'s
and select the most appropriate $K$. The cross-validation
splitting ratio for each sub-dataset will likely affect the selection
of $K$. We recommend using half-half splitting for each sub-dataset. Because the sample size $n_{i}$, the modeling methods $\mathcal{M}_{i}$, and the existence of data
heterogeneity are different across learners, it will be nontrivial and interesting to study how to decide the splitting ratios of cross-validation. We leave that as future work and refer interested readers to \citep{DingOverview,TargetCV} for related discussions on cross-validation.}

\section{Simulated Data Experiments \label{sec:Simulations}}

In this section, we present two simulation settings. Each example is repeated $100$ times. From a theoretical view, no standardization of the data is required since only the function
relationship between $Y$ and $\bm{X}$ matters. So one cluster may contain
two sub-datasets/learners whose responses or predictors are not on the same
scale. However, the nonparametric method usually requires
compact support, which may cause some computational issues. In the experiments, we standardize $\bm{x}$ and $y$ in each sub-dataset/learner.

\subsection{Simulation 1: clustering accuracy}

\begin{figure}[tb]
\centering \includegraphics[width=0.5\linewidth]{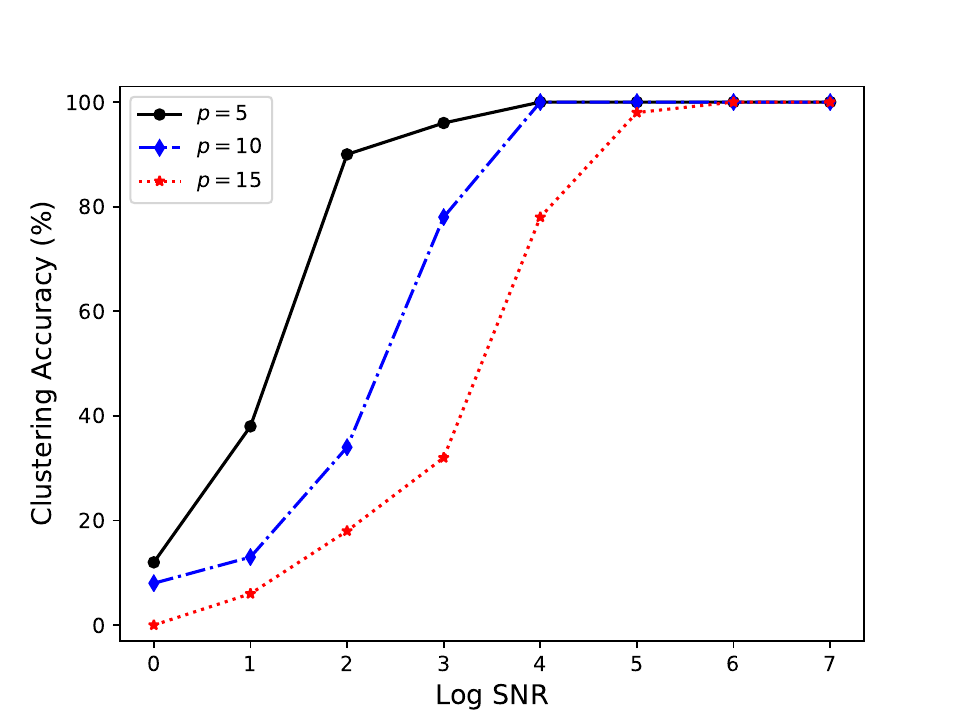}
\vspace{-0.1in}
\caption{{\small Clustering accuracy of the SEC algorithm for Simulation 1.\label{fig_A1} }}
\end{figure}

\begin{figure}[tb]
\begin{subfigure}{.48\textwidth} \centering 
 \includegraphics[width=1\linewidth]{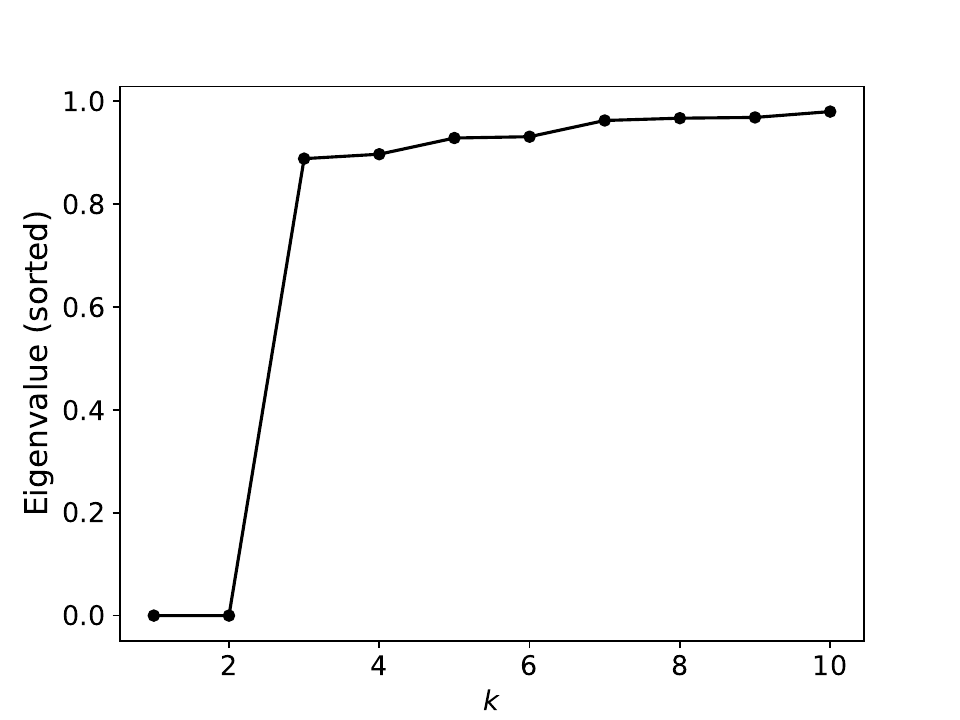} \caption{{\small Eigenvalues used to determine the gap statistic and number of clusters.}}
\label{fig_A2} \end{subfigure} \begin{subfigure}{.48\textwidth}
\centering 
 \includegraphics[width=1\linewidth]{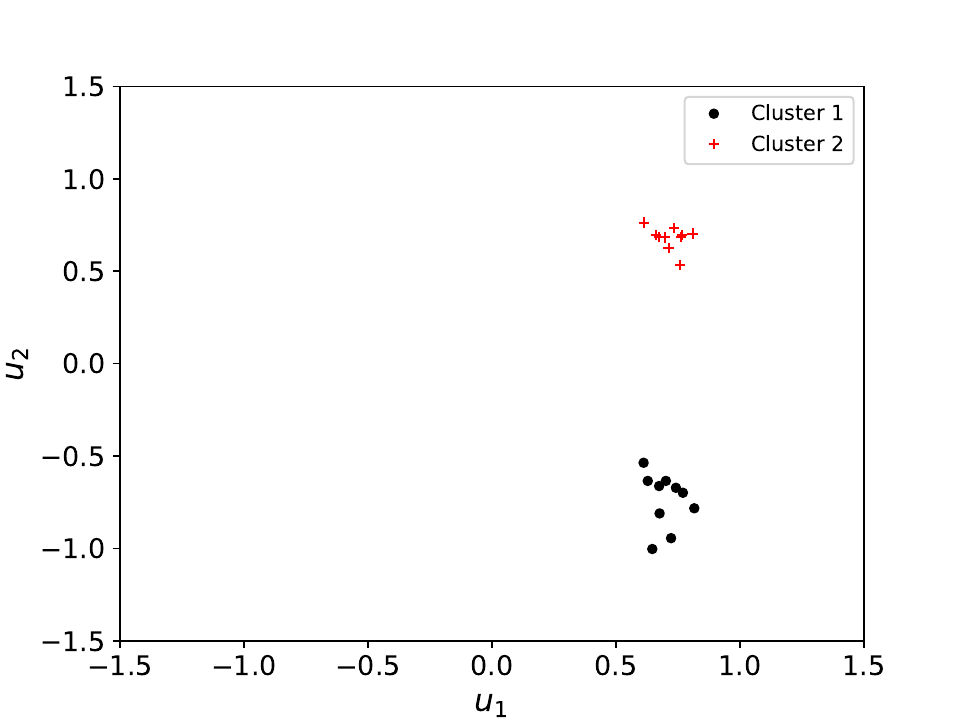} \caption{{\small Eigenvectors of the learners, indicated by the underlying true labels.}}
\label{fig_A3} \end{subfigure} \caption{{\small An illustration of the clustering results for Simulation 1, based on a realization
with $p=5$, SNR=$2^4$.}}
\label{fig_A23} 
\end{figure}

This example is to demonstrate the clustering accuracy of our
method. A clustering result is accurate if the number
of clusters is accurately identified, and each learner's label matches
the underlying truth (up to a permutation).
Suppose there are $20$ learners, $\{l_{i}\}_{i=1}^{20}$, 
each with a sub-dataset $D_{i}$ containing $n_{i}=50$ observations
and $p=5,10,20$ predictors.
The data of the first ten learners are generated
from the underlying model $Y=f_{1}(  \bm{X})+\v_{1}=\bm{\beta}_{1}^{\T}  \bm{X}+\v_{1}$,
where $  \bm{X}\sim \mathcal{N}(0,\mathtt{I}_p)$, $\v_{1}\sim \mathcal{N}(0,\sigma^{2})$, and $\bm{\beta}_{1}\in\mathbb{R}^{p}$.
The data of the remaining 10 learners are generated from $Y=f_{2}(  \bm{X})+\v_{2}=\bm{\beta}_{2}^{\T}  \bm{X}+\v_{2}$,
with $\v_{2}\sim \mathcal{N}(0,\sigma^{2})$, and $\bm{\beta}_{2}\in\mathbb{R}^{p}$.
We randomly generate $\bm{\beta}_{1}$ and $\bm{\beta}_{2}$ from the standard
Gaussian distribution (both $\bm{\beta}_{1}$ and $\bm{\beta}_{2}$ are set
as fixed in each replicated experiment such that $\bm{\beta}_{1}\neq\bm{\beta}_{2}$).
The signal-to-noise ratio (SNR) is defined by $\E(\|\bm{\beta}\|^{2})/\E(\v^{2})$,
which reduces to $p^{2}/\sigma^{2}$ in this case. We set the SNR level
to be one of the following: $2^{0},\ldots,2^{7}$, and the corresponding noise level $\sigma^2=p^2/\textnormal{SNR}$ falls into the range of $25/128$ to $400$. 
In the SEC algorithm, let each learner consider two candidate methods:
LASSO~\citep{10.2307/2346178}, with built-in half-half cross-validation
to select the tuning parameter, and Random Forest, with 50 trees and
depth 3. We apply the SEC algorithm to cluster the 20 sub-datasets. The
averaged clustering accuracy over $100$ replications is presented in
Figure~\ref{fig_A1}. We can see that the clustering accuracy
increases as the SNR increases. \REV{Also, for a fixed SNR, a smaller $p$
tends to lead to better clustering accuracy. It is mainly because a
less parsimonious model suffers from more estimation variance} given
the same amount of data. We also see that for a fixed $p$, the accuracy
curve tends to be flat when SNR is larger than $2^{5}$, showing the
SEC algorithm's robustness against high noise levels. In Figure \ref{fig_A23}, we also
present the result of a replication of the simulation with $p=5$
and SNR$=2^{4}$, \REV{with clustering accuracy near $100\%
$.} The eigenvalues used to apply the gap statistic are plotted in Figure \ref{fig_A2}. The eigenvectors in the spectral
clustering algorithm are shown in Figure \ref{fig_A3}.

\subsection{Simulation 2: robustness against candidate models}

In this example, we demonstrate that our method is robust against
candidate models in the cross-validation part of the ``select'' step.
Suppose there are $20$ learners, $\{l_{i}\}_{i=1}^{20}$, 
each with a sub-dataset $D_{i}$ containing $n_{i}=100$ observations
and $p=500$ predictors. We use the following two benchmark datasets
described in \citep{friedman1991multivariate,breiman1996bagging}.
The sub-datasets of the first ten learners are generated from 
$
Y=f_{1}(  \bm{X})+\v_{1}=\sqrt{X_{1}^{2}+(X_{2}X_{3}-1/(X_{2}X_{4}))^{2}}+\v_{1}, 
$
and the sub-datasets of the remaining ten learners are generated from
$
Y=f_{2}(  \bm{X})+\v_{2}=\arctan(X_{2}X_{3}-1/(X_{2}X_{4})/X_{1}+\v_{2},
$
where $X_{1}\sim U(0,100)$, $X_{2}\sim U(40\pi,560\pi)$, $X_{3}\sim U(0,1)$,
$X_{4}\sim U(1,11)$, and $\v_{1},\v_{2}\sim \mathcal{N}(0,0.01)$ are independent.
The remaining 496 predictors $\{X_{5},\ldots,X_{500}\}$ follow a standard
multivariate gaussian distribution $\mathcal{N}(0,\mathtt{I}_{496})$.

For each learner, we consider the candidate methods: Random
Forest (RF), \textit{k}-nearest neighbors (KNN), Support Vector Regression (SVR) \citep{drucker1997support},
Neural Network (NN), Gradient Boosting \citep{friedman2001} (GB),
LASSO, Least Angler Regression \citep{efron2004least}(LARS), Elastic
Net\citep{Zou05regularizationand} (EN), Ridge Regression (Ridge). 
To show the robustness of our procedure against
the number of candidate models and against the types of candidate
models, we consider four different choices of $\M_{i}$: \{RF, KNN,
SVR, NN, GB, LASSO, LARS, EN, Ridge\}, $\{\textrm{RF}, \textrm{KNN}, \textrm{SVR}, \textrm{GB}, \textrm{LASSO}, \allowbreak \textrm{LARS}, \textrm{EN}\}$,
$\{\textrm{RF, KNN, GB, LASSO, LARS}\}$, $\{\textrm{RF, GB, LASSO}\}$,
and $\{\textrm{GB}\}$. The results are presented in Table \ref{tab:robustness_against_candid}.
The clustering accuracy is stable over different choices of $\M_{i}$.
We can see the robustness of our method against both the number of
candidate models and the type of candidate models.
\begin{table}[tb]
\small
\begin{centering}
\begin{tabular}{c|ccc|cc}
\multirow{2}{*}{$|\mathcal{M}{}_{i}|$} & Proportion being selected & \multirow{2}{*}{Accuracy} & \multirow{2}{*}{$\hat{K}$} & Collaboration & No collaboration\tabularnewline
 & $\textrm{ }\textrm{(GB, RF, LASSO) \ }$ &  &  & MSE & MSE\tabularnewline
\hline 
1 & (1, 0, 0) & 66.0 & 2 & 0.100(0.0056) & 0.133(0.0038)\tabularnewline
3 & (0.56, 0.11, 0.33) & 74.0 & 2 & 0.095(0.0053) & 0.131(0.0042)\tabularnewline
5 & (0.56, 0.10, 0.34) & 58.0 & 2 & 0.087(0.0049) & 0.125(0.0044)\tabularnewline
7 & (0.57, 0.11,0.32) & 70.0 & 2 & 0.096(0.0056) & 0.134(0.0051)\tabularnewline
9 & (0.57, 0.10, 0.33) & 64.0 & 2 & 0.060(0.0018) & 0.112(0.0034)\tabularnewline
\end{tabular}
\par\end{centering}
\caption{\REV{{\small Prediction performance with collaboration and without collaboration, under various sets of candidate methods (rows). The column ``Proportion
being selected'' is the proportion of each method being selected
as the best method. The column ``Accuracy'' is the clustering accuracy
of the SEC algorithm. The standard
error of the averaged MSE over $100$ replications is reported in parentheses. The $\hat{K}$ denotes
the estimated number of clusters.\label{tab:robustness_against_candid}}}}
\end{table}

Without loss of generality, we focus on the first learner $l_{1}$ to evaluate whether the SEC algorithm improves prediction accuracy. We generate
a test set $\D_{\textrm{test}}=\{(y_{i}^{\textrm{test}},  \bm{x}_{i}^{\textrm{test}})\}^{100}_{i=1}$ generated
from the model $Y=f_{1}(  \bm{X})+\v_{1}$.  We consider two
modeling methods: No collaboration and Collaboration. The ``Collaboration''
method first applies the SEC algorithm and identifies learners in the same cluster as $l_{1}$. Then, we obtain the prediction
for the test set $\D_{\textrm{test}}$ based on the simple average of the estimated
predictors from those learners, as described in the formula \eqref{eq:futurepred}. The ``No Collaboration''
method simply fits $l_{1}$'s favored method on its own sub-dataset $\D_{1}$ and
applies the estimator on the test set $\D_{\textrm{test}}$ to make predictions.
The mean squared errors of the above two methods' predictions
are also shown in Table \ref{tab:robustness_against_candid}. Overall, ``Collaboration'' has a smaller MSE than ``No Collaboration.'' For $|\mathcal{M}_i|=1$, a right-sided \textit{t}-test of the MSE's of ``No Collaboration'' to that of ``Collaboration'' produces a \textit{p}-value of $2.1\times 10^{-6}$. We also observe significantly small \textit{p}-values for other cases of $|\mathcal{M}_i|$. When the number of candidate models in $\M_{i}$ is larger, the MSE of the ``Collaboration''
method is smaller. The above is because more candidate models
in the cross-validation part of the ``select'' step enable us to understand better the function relationship between the response and the predictors
so that the similarity matrix can better capture the true underlying
clusters. The prediction accuracy of the two methods is also stable across different
choices of $\M_{i}$, in terms of both the size of $\M_{i}$, $|\M_{i}|$ and the methods in $\M_{i}$.

\section{Real Data Applications\label{sec:Real-Data-Applications}}
In this section, we apply the SEC algorithm in two real data examples. 
\subsection{Application 1: CT Image Data}\label{sec:61}
We investigate the CT Image dataset in \citep{graf20112d} that consists
of $53500$ CT slices and $385$ variables. These $53500$ CT slices are obtained
from $97$ CT scans, where $74$ patients ($43$ male and $31$ female) took at
most a thorax scan and a neck scan. The response variable
is the relative location of the CT slice on the axial axis. The relative location of the CT slice on the axial axis is critical for registering CT scans in a body atlas \citep{graf20112d}, which enables the comparison of different CT scans. This dataset
has a natural sub-dataset structure since many CT slices are from
the same CT scan that can be treated as a sub-dataset.

We divide the dataset into $97$ sub-dataset/learners, each containing all the CT slices from a single CT scan.
Our goal is to find any potential clustering
structure (and the corresponding variable) that improves both scientific
understanding and predictive performance. We randomly divide these
$97$ learners into two parts: the training set ($64$ learners) and the
test set ($33$ learners). Similar to the data fairness example, for
each of the $33$ learners in the test set, we divide the sub-dataset
into two sets of equal size.

For the training set, we consider \REV{three} methods: ``clustering (pooled)'', \REV{``clustering (unpooled)''}, and ``no clustering''.
The ``no clustering'' method directly trains a Random Forest model
on the training set. The ``clustering (pooled)'' method first applies the
SEC algorithm to classify the learner in the training set into clusters, with $\M_{i}=\{\textrm{RF},\textrm{LASSO}\}$
for $i=1,\ldots,64$. Then it trains a Random Forest model separately
in each identified cluster (with all the within-cluster sub-datasets pooled). In contrast, the ``clustering (unpooled)'' does not pool the sub-datasets in the cluster but trains a Random Forest in each sub-dataset. For sub-datasets/learners in the validation set, the `no clustering' method directly applies the trained random
forest model to all the learners and obtains the overall mean squared error. The ``clustering (pooled)'' method first determines to cluster each learner belongs to and then applies the cluster-level trained random forest model. \REV{In contrast, the ``clustering (unpooled)'' method applied a weighted average as in equation \eqref{eq:futurepred}.}
\REV{
We repeat the data splitting $100$ times and summarize the results in
Table~\ref{tab:CT}. The results show that both options (unpooled and pooled) can significantly outperform
that of the ``no clustering'' method. A right-sided paired $\textit{t}$-test
that compares the MSE of ``clustering (unpooled)'' and ``clustering (pooled)'' with that of ``no clustering'' produces \textit{p}-values of $1.43 \times 10^{-13}$ and $1.77 \times 10^{-14}$,
respectively. The ``clustering
(unpooled)'' improves the MSE by 31$\%$ than ``no clustering'', and the ``clustering (unpooled)'' has
a slightly worse MSE compared with ``clustering (pooled)''. This
demonstrates the promising performance of collaborative learning even without pooling data.}
\begin{table}[tb]
\begin{centering}
\begin{tabular}{c|ccc}
 & Clustering (pooled) & Clustering (unpooled) & No Clustering\tabularnewline
\hline 
MSE & 94.09 (4.18) & 103.07(3.79) & 150.52 (2.10)\tabularnewline
$\hat{K}$ & \multicolumn{2}{c}{2 (6 times) and 3 (94 times)} & N/A\tabularnewline
\end{tabular}
\par\end{centering}
\caption{
\REV{{\small Results for the CT Image Data. The value in the parenthesis is the
standard error of the averaged MSE over $100$ replications, and $\hat{K}$
denotes the estimated clusters.}\label{tab:CT}}}
\end{table}
We also looked for a scientific understanding of the identified
clusters on the training set. So we investigated possible variables related to the cluster structure discovered by the SEC. Unfortunately, either the gender of the patient or whether the CT scan is from
the thorax or neck is not available in the dataset \citep{graf20112d}. However, this example does show the possibility of finding essential variables related to the cluster structure if further information is provided. Additionally, we can significantly improve the predictive performance without assessing any patient private information but the CT images themselves.

\subsection{Application 2: Electrical Grid Stability Data}

This example is to demonstrate the performance of the SEC algorithm
when the data are under adversarial attacks. The Electrical Grid Stability
Data \citep{8587498} consists of 10000 observations and $14$ variables.
Among the $14$ variables, two variables describe the system stability:
one is categorical (stable/unstable), and the other is continuous (a positive value means a linearly unstable system). We use the continuous
variable as the response. The other $12$ variables are the input
of the \textit{Decentral Smart Grid Control} system.

We first divide the data into training set ($n_{1}=8000$) and the
test set ($n_{2}=2000$). The training set is randomly divided into
$50$ learners, each with $160$ observations. We may assume the data
are stored in $50$ servers, and some servers get attacked by hackers. \REV{Let $d=0,1,\ldots,49$ denote the number of attacked learners. We set that the first $d$ out of $50$ learners are attacked.} Each time a sub-dataset is ``attacked,''
we change the response variable to the negative of its original
value. We also assume that the $50$th learner knows that its dataset is not attacked.

\begin{figure}[tb]
\centering{}\includegraphics[width=0.55\linewidth]{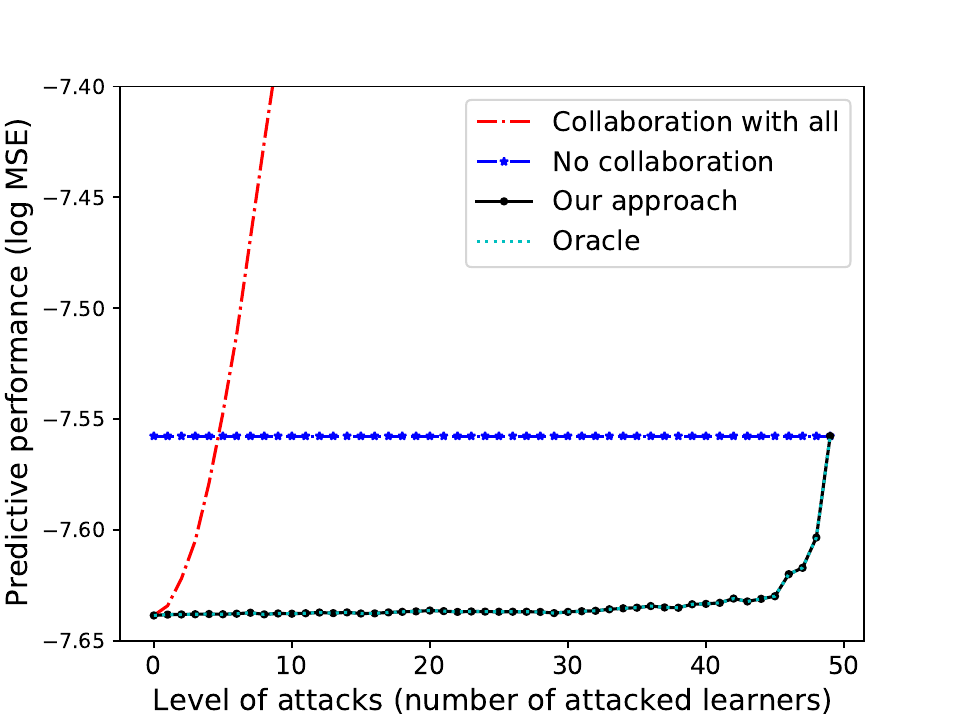}
\vspace{-0.1in}
\caption{{\small Prediction error (evaluated by MSE) as an increasing function of attack
severity.\label{fig_demo2} }}
\end{figure}
Under potential attacks, we consider four options of the 50th learner
to perform data analysis, denoted as ``Collaboration with all'', ``No collaboration'',
``Our approach'', and ``Oracle''. The   ``Collaboration
with all'' option ignores the fact that some learners/sub-datasets
are attacked and insist on collaborating with all the other learners.
In the ``No collaboration'' option, a learner (say the 50th) trusts nobody but itself
and uses its sub-dataset for learning. In the ``Our approach'' option, the 50 learners are clustered by SEC into ``attacked''
and ``intact''. Then the learners classified
as intact will collaborate. The ``Oracle'' option means that an oracle knows
which learners are attacked and collaborates with those intact ones.
In collaboration, we allow the learners
to share datasets. In other words, once a learner identifies collaborators, the learner pools the data and fits a linear regression.

The trained linear model is then applied to the test set to evaluate
its performance (MSE). We plot the predictive performance against the
number of attacked learners in Figure \ref{fig_demo2}. \REV{We only present part of the red curve since it explodes as the level of attacks increases. The value of the red curve increases from $-7.65$ to $-5.25$ when the number of attacked learners increases from $0$ to $49.$} As the proposed method accurately clusters all the intact learners,
the performance curve of ``Our approach''
overlaps with that of ``Oracle''. We also see that the predictive
performance of ``Our approach'' decreases when
the level of attack (meaning the number of the attacked learners) increases.
In particular, the decrease becomes very sharp when the number of
attacked learners is greater than $45$. One reason is that the linear
model based on the information of one sub-dataset (with a sample size of
$160$ and $12$ predictors) or two is enough to capture the underlying
relationship. Indeed, the scale of MSE is very small ($10^{-4}$). So collaborating with more than five intact learners may not
improve the prediction accuracy much compared with collaborating
with only two intact learners.

\REV{The proposed SEC algorithm
can be applied even though each learner can only access its own sub-dataset. Nevertheless, the SEC algorithm can be applied when each learner has access to all the sub-datasets. In such cases, we envision it as a pre-screening
method to screen out contaminated sub-datasets, which improves modeling and prediction accuracy.} 

\section{Conclusion}
\label{sec:con}
This paper proposed a framework of \textit{meta clustering} for selecting ``qualified'' collaborators for collaborative learning. 
If two datasets exhibit a similar underlying relationship between the response and predictors, they fall in the same cluster. We developed a clustering algorithm named SEC to perform meta clustering efficiently. It only requires the exchange of fitted functions instead of raw data to evaluate the similarity among datasets. 
We showed promising applications of the framework to enhance data fairness, improve single-learner prediction accuracy, and discover potential grouping structures of a dataset.  

\subsection*{Appendix}
\appendix
\section{Technical proofs}
\subsection{Notation}
For a sequence of random variables $\{X_{n}\}$ and a deterministic
sequence $\{a_{n}\}$, $n=1,2,\ldots$, we write $X_{n}\lesssim a_{n}$
if $X_{n}=O_{p}(a_{n})$. We write $X_{n}\gtrsim a_{n}$ if for any constant
$\delta\in(0,1)$ there exists a $c_{\delta}>0$ such that $\liminf_{n\rightarrow\infty}\P(X_{n}\geq c_{\delta}a_{n})\geq1-\delta$. If both $X_{n}\lesssim a_{n}$ and $X_{n}\gtrsim a_{n}$, we then write $X_{n}\sim a_{n}$.

An estimator $\{\hat{f}_{n}\}_{n=1}^{\infty}$ is said to converge
in probability to $f$ at the rate $\{a_{n}\}$ if $\norm{\hat{f}_{n}-f}_{2}\sim a_{n}$.
We use the notation $\delta_f$ to denote the modeling method that corresponds to the data generating function $f$. For example, let $f(\bm{x})=x_1+x_2+x_3$ and then we can denote the linear regression of the response on the variables $\{x_1, x_2, x_3\}$ as $\delta_f$.

\subsection{Assumptions}
\begin{assumption}\label{ass3} The number of learners $L$ is a
fixed positive integer. For each $n_{i}$, $i=1,\ldots,L$, $n_{i}\sim n$
as $n\rightarrow\infty$ where $n=n_{1}+\cdots+n_{L}$. \end{assumption}

\begin{assumption}\label{ass2} For each $f_{i}$ and its estimator
$\hat{\delta}_{i}\in\M_{i}$ from $\D_{i}$, $\exists\ M_{1}>0$ such
that $||f_{i}(x)-\hat{f}_{i}(x)||_{\infty}<M_{1}$, $\forall\ n\geq1$.
\end{assumption}

\begin{assumption}\label{ass_nonparExist} We assume that each learner
uses at least one nonparametric modeling method: $\Mn_{i}\neq\emptyset$,
$\forall\ i=1,\ldots,L$. \end{assumption}

\begin{assumption}\label{ass1} There exists sequences $\{\ell_{n}\}$
and $\{u_{n}\}$ such that 
\begin{align}
 & \sqrt{n}\min\{\ell_{n},u_{n}\}\rightarrow\infty\textrm{ as }n\rightarrow\infty.\label{eq_nonDOWN}\\
 & \max\{\ell_{n},u_{n}\}\rightarrow0\textrm{ as }n\rightarrow\infty,\label{eq_nonUP}
\end{align}
and that for each $i=1,\ldots,L$, for any nonparametric estimator
$\hat{f}_{n_{i}}\in\Mn_{i}$, 
\begin{align}
 & \ell_{n_{i}}\lesssim\norm{f_{i}-\hat{f}_{n_{i}}}_{2}\lesssim u_{n_{i}}.\label{eq_non}
\end{align}
Moreover, for each $i=1,\ldots,L$, for any parametric estimator $\hat{f}_{n_{i}}=f_{\hat{\theta}_{n_{i}}}\in\Mp_{i}$,
\begin{align}
\norm{f_{i}-\hat{f}_{n_{i}}}_{2}\sim n_{i}^{-1/2}\qquad & \textrm{ if }f_{i}\in\{f_{\theta}:\theta\in\Theta_{i}\}\label{eq_para_well}\\
\norm{f_{i}-\hat{f}_{n_{i}}}_{2}\sim1\qquad & \textrm{ if }f_{i}\not\in\{f_{\theta}:\theta\in\Theta_{i}\}.\label{eq_para_mis}
\end{align}
\end{assumption} \begin{assumption}\label{ass_BoundedDifference}
The difference between any two functions $\Delta f_{i,j}=f_{j}-f_{i}$
is bounded, i.e., $||\Delta f_{i,j}||_{\infty}<\infty$
\end{assumption}




\begin{assumption}\label{ass_BoundedCondition-3}The underlying functions
$f^{(i)}$ are bounded, i.e., $||f^{(i)}||_{\infty}<\infty$.\end{assumption}


Assumption~\ref{ass1} says that all the nonparametric methods
will be consistent in estimating the underlying regression function (in
the sense of $L_{2}$ distance). When the model class $\Mp$ includes
the data generating regression function, namely the data generating model is parametric, the correct parametric models will converge to the truth
at a parametric (which is the square root of sample size), faster than
nonparametric rates. However, when $\Mp$ does not include the underlying regression
function, then the estimation error will be bounded away from zero
in probability. This assumption is very mild and common in regression
analysis. More detailed discussions could be found in, e.g., \citep{lu2019assessing}
and the reference therein.

\subsection{Proof of Theorem 4} 
To prove the theorem, we first prove the following three lemmas, which provide theoretical guarantees of the SEC algorithm's three steps.

\subsubsection{Step 1}
In the following lemma we show that the cross-validation in Step 1 will select the correct one if the underlying function is parametric and specified in the candidate methods. Otherwise, nonparametric methods are favored.
\begin{lemma}
\label{thm_par_nonpar_distinct}Assume Assumptions \ref{ass2}, \ref{ass_nonparExist} and \ref{ass1} hold. For each learner $l_{i}$, if the corresponding modeling method of the underlying data generating function is specified in the candidate models, namely $\delta_{f_{i}}\in\M_{i}^{p}$, then the $\delta_{f_{i}}$ will be selected by the cross-validation in step 1 with probability going to one as $n\rightarrow\infty$. Otherwise, there exists a method $\delta^{\prime}\in\M_{i}^{non}$ that minimizes the cross-validation error (equation (2) in the article).
\end{lemma}
\begin{proof}
For any candidate modeling method $\delta\in\M_{i}$, denote its corresponding estimator as $\hat{\delta}_{n_{i}/2}$ where the subscript denotes that the estimator is obtained from the training set with sample size $n_{i}/2$.  The cross-validation error of the method $\delta$ is 
\begin{align*}
e(\hat{\delta}_{n_{i}/2})	:&=\frac{2}{n_{i}}\sum_{t=n_{i}/2+1}^{n_{i}}\bigl(f_{i}(\bm{x}_{t})+\v_{it}-\hat{\delta}_{n_{i}/2}(\bm{x}_{t})\bigr)^{2}\\
	&=\frac{2}{n_{i}}\sum_{t=n_{i}/2+1}^{n_{i}}\bigl(f_{i}(\bm{x}_{t})-\hat{\delta}_{n_{i}/2}(\bm{x}_{t})\bigr)^{2}+\frac{2}{n_{i}}\sum_{t=n_{i}/2+1}^{n_{i}}\v_{it}^{2} \\
	&\quad+\frac{4}{n_{i}}\sum_{t=n_{i}/2+1}^{n_{i}}\bigl(f_{i}(\bm{x}_{t})-\hat{\delta}_{n_{i}/2}(\bm{x}_{t})\bigr)\v_{it} .
\end{align*}
For a sequence $\{X_n\}$ where each element has a finite variance, we have $X_n-E(X_n)=O_p(\sqrt{\var(X_n)})$. So we have $\frac{2}{n_{i}}\sum_{t=n_{i}/2+1}^{n_{i}}\v_{it}^{2}=\sigma_{i}^{2}+O_{p}(n_{i}^{-1/2})$. By Assumption \ref{ass2}, we have $$\frac{4}{n_{i}}\sum_{t=n_{i}/2+1}^{n_{i}}\bigl(f_{i}(\bm{x}_{t})-\hat{\delta}_{n_{i}/2}(\bm{x}_{t})\bigr)\v_{it}=O_{p}(n_{i}^{-1/2})$$and $$\frac{2}{n_{i}}\sum_{t=n_{i}/2+1}^{n_{i}}\bigl(f_{i}(\bm{x}_{t})-\hat{\delta}_{n_{i}/2}(\bm{x}_{t})\bigr)^{2}=\norm{f-\hat{\delta}_{n_{i}/2}}^{2}+O_{p}(n_{i}^{-1/2}).$$

Together with Assumption \ref{ass1}, we have $e(\hat{\delta}_{n_{i}/2})=\norm{f-\hat{\delta}_{n_{i}/2}}^{2}+\sigma_{i}^{2}+O_{p}(n_{i}^{-1/2})\sim\sigma_{i}^{2}+O_{p}(\max\{n_{i}^{-1/2},u_{n_{i}}^{2}\})$.

If $f_{i}$ is a parametric model and the corresponding modeling method is contained in $\M_{i}^{p}$, then by Assumption \ref{ass1}, $$\norm{f-\hat{\delta}_{n_{i}/2}}^{2}+\sigma_{i}^{2}+O_{p}(n_{i}^{-1/2})\sim\sigma_{i}^{2}+O_{p}(n_{i}^{-1/2})$$
for a parametric $\hat{\delta}_{i}$ and 
\[
\norm{f-\hat{\delta}_{n_{i}/2}}^{2}+\sigma_{i}^{2}+O_{p}(n_{i}^{-1/2})\sim\sigma_{i}^{2}+O_{p}(\max\{n_{i}^{-1/2},u_{n_{i}}^{2}\})
\]
for a nonparametric $\hat{\delta}_{i}$. So if the $u_{n}\succ n^{-1/4}$, the selection process will pick the correct parametric model
$\delta$.

If $f_{i}$ is a parametric model and the corresponding modeling method
is not contained in $\M_{i}^{p}$, or $f_{i}$ is a nonparametric
model, we have $\norm{f-\hat{\delta}_{n_{i}/2}}^{2}+\sigma_{i}^{2}+O_{p}(n_{i}^{-1/2})\sim\sigma_{i}^{2}+1$
for parametric $\delta$ and $\sim\sigma_{i}^{2}+O_{p}(\max\{n_{i}^{-1/2},u_{n_{i}}^{2}\})$
for a nonparametric $\delta$. Thus, nonparametric models will be favored. 
\end{proof}

\subsubsection{Step 2} The following lemma states that the dissimilarity $v_{ij}$ constructed in the Exchange step has a well-separation property.
\begin{lemma}
\label{thm_exchange_validation} Assume Assumptions  \ref{ass2}, \ref{ass_nonparExist}, \ref{ass1},and \ref{ass_BoundedDifference} hold. If $i,j$ are in the same cluster, namely $f_{i}=f_{j}$, we have $v_{ij}=o_{p}(1)$ as $n\rightarrow\infty$. Otherwise, we have $v_{ij}=||f_{i}-f_{j}||_{2}^{2} + ||\Delta f_{i,j}||_1E||f_{j}-\hat{f}_{n_{i}}||_1+o_{p}(1)$, namely bounded away from 0 in probability.

\end{lemma}
\begin{proof}
Suppose the fitted model $\hat{f}_{n_{i}}$ from learner $i$ is applied
to data $\D_{j}$. Let $\E_{n_{j}}g=n_{j}^{-1}\sum_{t=1}^{n_{j}}g(\bm{x}_{t})$
be the empirical expectation for any measurable function $g(\cdot)$.
Let $\Delta f_{i,j}=f_{j}-f_{i}$. Then the quadratic loss $\hat{e}_{i\rightarrow j}$
satisfies 
\begin{align*}
\hat{e}_{i\rightarrow j} & =\Ej\bigl(f_{j}-\hat{f}_{n_{i}}+\v_{j}\bigr)^{2}\\
 & =\E_{n_{j}}\bigl(\Delta f_{i,j}\bigr)^{2}+\Ej\v_{j}^{2}+\Ej\bigl(f_{i}-\hat{f}_{n_{i}}\bigr)^{2}+2\Ej\{\Delta f_{i,j}\bigl(f_{i}-\hat{f}_{n_{i}}\bigr)\}+2\Ej\bigl(f_{j}-\hat{f}_{n_{i}}\bigr)\cdot\v_{j} \\
 &\quad+2\Ej\Delta f_{i,j}\cdot\v_{j}.
\end{align*}
By Assumptions \ref{ass2} and \ref{ass_BoundedDifference}, we have 
$$2\Ej\{\Delta f_{i,j}\bigl(f_{i}-\hat{f}_{n_{i}}\bigr)\}+2\Ej\bigl(f_{j}-\hat{f}_{n_{i}}\bigr)\cdot\v_{j}+2\Ej\Delta f_{i,j}\cdot\v_{j}=||\Delta f_{i,j}||_1E||f_{j}-\hat{f}_{n_{i}}||_1+O_p(n_{j}^{-1/2}).$$ 
By Assumption~\ref{ass_nonparExist}, $\hat{f}_{j}$ is at least
consistent in estimating $f_{j}$ at a nonparametric rate. Thus
we have 
\begin{align}
\hat{e}_{i\rightarrow j}=||\Delta f_{i,j}||_2^{2}+\Ej\v_{j}^{2}+O_{p}(u_{n_{i}}^{2})+O_{p}(n_{j}^{-1/2}).\label{eq_eij}
\end{align}
Similar arguments lead to 
\begin{align}
\hat{e}_{j}=\Ej\v_{j}^{2}+\Delta f_{i,j}E||f_{j}-\hat{f}_{n_{i}}||_1+O_{p}(u_{n_{j}}^{2})+O_{p}(n_{j}^{-1/2}).\label{eq_ej}
\end{align}
Therefore, $\hat{e}_{i\rightarrow j}-\hat{e}_{j}=o_{p}(1)$ if $f_{i}=f_{j}$,
and otherwise $v_{ij}=||f_{i}-f_{j}||_{2}^{2}+||\Delta f_{i,j}||_1E||f_{j}-\hat{f}_{n_{i}}||_1+o_{p}(1)$. 

\end{proof}

\begin{rem}
Theoretically
speaking, when the sample size is large enough, any reasonable clustering
algorithm works. We use the spectral clustering algorithm stated in Algorithm 1 for practical constraints such
as sample size and data heterogeneity.
\end{rem}
\subsubsection{Step 3}
We show in the following lemma that the spectral clustering algorithm can accurately identify the $K$ clusters when the sample size $n_{i}$ goes to infinity for each learner $l_i$.
\begin{lemma}\label{lemma2}
\label{cor_classification}  Assume Assumptions A1-A4 in Appendix \ref{proof} hold. Let $u_{j}^{(i)}$ denotes the  $j$-th
row of the  $i$-th subgroup of $ \mathtt{U}_{*}$ in the spectral clustering
algorithm. There exist $K$ orthonormal vectors $r_{1},\ldots,r_{K}$
such that the rows of $ \mathtt{U}_{*}$ satisfy 
\[
\frac{1}{n}\sum_{t=1}^{K}\sum_{j=1}^{|\mathcal{S}_t|}||u_{j}^{(t)}-r_{t}||_{2}^{2}\le4C(4+2\sqrt{K})^{2}\frac{\epsilon^{2}}{(\delta-\sqrt{2}\epsilon)^{2}},\label{eq_lemma3}
\]
for some positive constants $\epsilon,\delta,C$. In addition, when $K<\infty$ is unknown, if the penalty term satisfies that $O_{p}(\max\{n_{i}^{-1},u_{n_{i}}^{4}\})=o(\lambda_{n})$ and $\lambda_{n}=o(1)$, then $K$ can be identified with probability going to one.
\end{lemma}
\begin{proof}
For notional convenience, the proof is deferred to Section \ref{proof} in the Appendix.
\end{proof}

\begin{rem}
In our meta learning framework, the above right-hand side is close to zero. To see this, we can take $\epsilon_{1},\epsilon_{2}\rightarrow0$
and $\delta$ fixed, so $\frac{\epsilon^{2}}{(\delta-\sqrt{2}\epsilon)^{2}}\sim\epsilon^{2}=K(K-1)\epsilon_{1}+K\epsilon_{2}^{2}\rightarrow0$
as the sample size $n_{i}\rightarrow\infty$ for any learner $i=1,\ldots,L$.
\end{rem}
\begin{rem}
The lemma indicates that the rows of $U_{*}$ will form tight
clusters around $K$ orthogonal points based on the $K$ true clusters.
Applying the \textit{k}-means clustering to $ \mathtt{U}_{*}$ will allow us to obtain accurate
clusters.
\end{rem}

\subsubsection{Summary} We conclude the proof of Theorem 4 from Lemma \ref{lemma2}. 

\subsection{\label{proof}Proof of Lemma \ref{cor_classification}}
This lemma applies \citep[Theorem 2,][]{ng2002spectral}. In the following, we show that the assumptions of \citep[Theorem 2,][]{ng2002spectral} are satisfied in our context. 

Denote the similarity matrix as $ \mathtt{S}=[ \mathtt{S}_{ij}]_{n\times n}$ with $ \mathtt{S}_{ij}$
representing the element in its  $i$-th row and  $j$-th column. Assume
the learners are ordered based on the cluster they are in, so the
elements in $ \mathtt{S}$ can be rearranged, and we denote the rearranged matrix
as 
\[ 
\left[\begin{array}{ccc}
 \mathtt{S}^{(11)} & \ldots &  \mathtt{S}^{(1k)}\\
\vdots & \ddots & \vdots\\
 \mathtt{S}^{(k1)} & \ldots &  \mathtt{S}^{(kk)}
\end{array}\right],
\]
with $ \mathtt{S}^{(tt)}$ denoting the similarity matrix between the learners
that belong to the $t$-th cluster, $t=1,\ldots,K$. We state below the
four assumptions and the resulting theorem in \citep{ng2002spectral}.
Let $\mathcal{S}_{t}$ denote the $t$-th cluster and $|\mathcal{S}_{t}|$ denotes the
cardinality of the $t$-th cluster.

\textbf{Assumption A1.} There exists $\delta>0$ such that $h_{t}^{2}/2\ge\delta$
for all $t=1,...,K$. Here, 
$$h_{t}:=\min_{I\subseteq\{1,\ldots,|S_{t}|\}}\frac{\sum_{j\in I,k\notin I} \mathtt{S}_{jk}^{(tt)}}{\min\{\sum_{j\in I}d_{j},\sum_{k\notin I}d_{k}\}}$$
is defined as the Cheeger constant of the cluster $\mathcal{S}_{t}$, where
$d_{j}:=\sum_{k} \mathtt{S}_{jk}^{(tt)}$ denotes the extent of the connectedness
for point $j$ to other points in $\mathcal{S}_{t}$.

\textbf{Assumption A2.} There is some fixed $\epsilon_{1}>0$, so
that for every $t_{1},t_{2}\in\{1,\ldots,K\}$ and $t_{1}\neq t_{2}$,
we have that 
\[
\sum_{j\in \mathcal{S}_{t_{1}}}\sum_{k\in \mathcal{S}_{t_{2}}}\frac{ \mathtt{S}_{jk}^{2}}{d_{j}d_{k}}\le\epsilon_{1}.
\]

\textbf{Assumption A3.} For some fixed $\epsilon_{2}>0$,for every
$t=1,\ldots,K$ and $j\in \mathcal{S}_{t}$, we have 
\[
\frac{\sum_{k:k\notin \mathcal{S}_{t}} \mathtt{S}_{jk}}{d_{j}}\le\epsilon_{2}(\sum_{k,l\in \mathcal{S}_{t}}\frac{ \mathtt{S}_{kl}^{2}}{d_{k}d_{l}})^{-1/2}.
\]

\textbf{Assumption A4.} There is some constant $C>0$ so that for
every $t=1,\ldots,K$ and $j\in \mathcal{S}_{t}$, we have $d_{j}^{(t)}\ge\frac{\sum_{k=1}^{|\mathcal{S}_{t}|}d_{k}^{(t)}}{C|\mathcal{S}_{t}|}$.
\begin{thm}\citep[Theorem 2,][]{ng2002spectral}
\label{thm:Let-Assumptions-A1-A4}Suppose that Assumptions A1-A4 hold. Set
$$\epsilon=\sqrt{K(K-1)\epsilon_{1}+K\epsilon_{2}^{2}}.$$ If $\delta>(2+\sqrt{2})\epsilon$,
 there exist $K$ orthonormal vectors $r_{1},\ldots,r_{K}$ such
that the rows of $ \mathtt{U}_{*}$ satisfy 
\[
\frac{1}{n}\sum_{t=1}^{K}\sum_{j=1}^{|\mathcal{S}_{t}|}||u_{j}^{(t)}-r_{t}||_{2}^{2}\le4C(4+2\sqrt{K})^{2}\frac{\epsilon^{2}}{(\delta-\sqrt{2}\epsilon)^{2}},
\]
where $u_{j}^{(t)}$ denotes the $j$-th row of the $t$-th sub-block
of $ \mathtt{U}_{*}$.
\end{thm}
\begin{proof}
We only need to check the four assumptions. Assumption A1 is satisfied
by $ \mathtt{S}_{jk}^{(tt)}>0$ for $j\neq k$. To check Assumption A2, for
any $j\in \mathcal{S}_{t_{1}},k\in \mathcal{S}_{t_{2}}$, we have $ \mathtt{S}_{jk}=o_{p}(1)$ if
$i_{1}\neq i_{2}$ and $ \mathtt{S}_{jk}\sim1$ if $i_{1}=i_{2}$. Since $K$
is fixed, we can find two positive constants $M_{1}$ and $M_{2}$
such that $M_{1}\le  \mathtt{S}_{jk}\le M_{2}$ for any $j\in \mathcal{S}_{t_{1}},k\in \mathcal{S}_{t_{2}},t_{1}\neq t_{2}$.
Thus, $d_{j}=\sum_{k} \mathtt{S}_{jk}^{(t_{1}t_{1})}\ge| \mathcal{S}_{t_{1}}|M_{1}$. When
$t_{1}\neq t_{2}$, we have
\[
\sum_{j\in \mathcal{S}_{t_{1}}}\sum_{k\in \mathcal{S}_{t_{2}}}\frac{ \mathtt{S}_{jk}^{2}}{d_{j}d_{k}}\le\sum_{j\in \mathcal{S}_{t_{1}}}\sum_{k\in \mathcal{S}_{t_{2}}}\frac{ \mathtt{S}_{jk}^{2}}{|\mathcal{S}_{t_{1}}||\mathcal{S}_{t_{2}}|M_{1}^{2}}\le\frac{(\max_{j\in \mathcal{S}_{t_{1}},k\in \mathcal{S}_{t_{2}}} \mathtt{S}_{jk})^{2}}{M_{1}^{2}}\sim(\frac{o_{p}(1)}{M_{1}})^{2}.
\]
Hence we can find a fixed $\epsilon_{1}>0$ such that $ (\max_{j\in \mathcal{S}_{t_{1}},k\in \mathcal{S}_{t_{2}}} \mathtt{S}_{jk})^{2} / M_{1}^{2} \le\epsilon_{1}$.
To check Assumption A3, when $t_{1}\neq t_{2}$, we have

\begin{align*}
\frac{\sum_{k:k\notin \mathcal{S}_{t_1}} \mathtt{S}_{jk}}{d_{j}}(\sum_{k,l\in \mathcal{S}_{t_1}}\frac{ \mathtt{S}_{kl}^{2}}{d_{k}d_{l}})^{1/2} & \le\frac{(L-|\mathcal{S}_{t_1}|)\cdot\max  \mathtt{S}_{jk}}{|\mathcal{S}_{t_1}|M_{1}}(\sum_{k,l\in \mathcal{S}_{t_1}}\frac{M_{2}^{2}}{|\mathcal{S}_{t_{1}}|^{2}M_{1}^{2}})^{1/2}\\
 & \le\frac{(L-|\mathcal{S}_{t_1}|)\max  \mathtt{S}_{jk}}{|\mathcal{S}_{t_1}|M_{1}}\frac{M_{2}}{M_{1}}\\
 & =\left(\max_{j\in \mathcal{S}_{t_{1}},k\in \mathcal{S}_{t_{2}}} \mathtt{S}_{jk}\right)\cdot(L/|\mathcal{S}_{t_1}|-1)\cdot\frac{M_{2}}{M_{1}^{2}}.
\end{align*}
Since $L$ is fixed, we have $\left(\max_{j\in \mathcal{S}_{t_{1}},k\in \mathcal{S}_{t_{2}}} \mathtt{S}_{jk}\right)\cdot(L/|\mathcal{S}_{t_1}|-1)\cdot\frac{M_{2}}{M_{1}^{2}}=o_{p}(1)$.
So we can find a fixed $\epsilon_{2}>0$ such that $\max_{j\in \mathcal{S}_{t_{1}},k\in \mathcal{S}_{t_{2}},t_{1}\neq t_{2}} \mathtt{S}_{jk}\cdot(L/|\mathcal{S}_{t_1}|-1)\cdot\frac{M_{2}}{M_{1}^{2}}\le\epsilon_{2}$.
To check Assumption A4, we have
\[
\frac{\sum_{k=1}^{|\mathcal{S}_{t}|}d_{k}^{(t)}}{d_{j}|\mathcal{S}_{t}|}\le\frac{|\mathcal{S}_{t}|M_{2}}{|\mathcal{S}_{t}|M_{1}|\mathcal{S}_{t}|}=\frac{M_{2}}{|\mathcal{S}_{t}|M_{1}}.
\]
We can find $C$ such that $\frac{M_{2}}{|\mathcal{S}_{t}|M_{1}}\le C$.

Thus all the assumptions are satisfied and Theorem \ref{thm:Let-Assumptions-A1-A4}
follows. So Lemma \ref{cor_classification} holds.

If $K$ is unknown, the penalized selection (equation (4) in the article) will lead to a $\hat{K}$ such that $\hat{K}\ge K)$ with high probability. To see this, let $ \mathtt{U}_*^K$, $ \mathtt{U}_*^{K+1}$, and $ \mathtt{U}_*^{K-1}$ be the matrices in step 2(b) of Algorithm 1 with the number of clusters being $K$, $K+1$ and $K-1$ respectively. From what we proved aforementioned, since the rows of $ \mathtt{U}_*^K$ will form tight clusters around $K$ orthogonal points, with assumption \ref{ass2}, it is not hard to see that 
$$\sum_{t=1}^{K}\sum_{i,j\in \mathcal{S}_{t}}\frac{1}{2|\mathcal{S}_{t}|}|| \mathtt{u}^{K}_{(i)}- \mathtt{u}^{K}_{(j)}||^{2}\sim o_p(1),$$
$$\sum_{t=1}^{K-1}\sum_{i,j\in \mathcal{S}_{t}}\frac{1}{2|\mathcal{S}_{t}|}|| \mathtt{u}^{K-1}_{(i)}- \mathtt{u}^{K-1}_{(j)}||^{2}\sim 2n,$$and
$$\sum_{t=1}^{K+1}\sum_{i,j\in \mathcal{S}_{t}}\frac{1}{2|\mathcal{S}_{t}|}|| \mathtt{u}^{K+1}_{(i)}- \mathtt{u}^{K+1}_{(j)}||^{2}\sim o_p(1),$$
since the rows of $ \mathtt{U}_*^{K-1}$ will not form tight clusters around neither $K$ or $K-1$ orthogonal points. It only remains to show that the design of $\lambda_n$ is gonna penalize picking a $\hat K$ that is larger than $K$. From Lemma \ref{thm_exchange_validation}, $v_{i,j}=O_{p}(\max\{n_{i}^{-1/2},u_{n_{i}}^{2}\})$, for $i,j$ being in the same clustering. So if two rows of $ \mathtt{U}_*$ is from the same cluster, then their Euclidean distance is of the same order as $v_{i,j}$. To successfully penalize ``dividing a `correct' (in the sense that elements in a cluster share the same data generating function) cluster with size $u_{1}+u_{2}$ into 
two clusters of size $u_{1},u_{2}$'', the penalty should be at least 
\[
O_{p}(v^2_{i,j})\cdot h(u_{1},u_{2}),
\]
where 
\[
h(u_{1},u_{2})=\frac{{u_{1}+u_{2} \choose 2}}{(u_1+u_2)}-\frac{{u_{1} \choose 2}}{u_1}-\frac{{u_{2} \choose 2}}{u_2}=1/2.
\]
Thus, the condition on $\lambda_{n}$ suffices.
\end{proof}

\section{Extended discussion on the statistical gain of our algorithm and the data privacy constraints}
\REV{
\paragraph{Statistical gain.}
Denote the best fitted model for learner $l_{i}$ as $\hat{f}_{i}$. For prediction, denote $\hat{y}_{c}(\bm{x}):=\sum_{i\in\mathcal{S}_{t}} w_{i} \hat{f}_{{i}}(\bm{x})$
as the weighted average of predictions from all local models in the
cluster $\mathcal{S}_{t}$, where $w_{i}:=n_{i}/\Sigma_{j\in\mathcal{S}_{t}}n_{j}$.
Intuitively, such a weighted predictor is expected to bring a statistical gain for learner $l_{1}$ when other learners in the same cluster, say $\mathcal{S}_{t}$, employ appropriate models. But when some learners fail to specify the most appropriate model to capture the underlying data generating mechanism, such a weighting scheme may not work well even compared with $l_{1}$'s local training. Let us consider a particular example.  Assume the underlying model for this cluster is
$\mathcal{S}_{t}$ is $y=h(x)+\epsilon$, where $\epsilon$ denotes the independent noise. We describe the statistical gain
of applying collaborative learning by $\mathcal{R}:=\mathbb{E}(y-\hat{f}_{1}(\bm{x}))^{2}-\mathbb{E}(y-\hat{y}_{c}(\bm{x}))^{2}=||h-\hat{f}_{1}||_{P_x}^{2}-||h-\sum_{i\in\mathcal{S}_{t}}w_{i}\hat{f}_{{i}}||_{P_x}^{2}$. 
If the data of learners in $\S_t$ are i.i.d., $h$ is a linear function with a fixed-dimensional parameter, and $h\in\mathcal{M}_{i}$ for all $i \in \S_t$, we have a gain, namely $\mathcal{R}>0$. This is because the weighting can reduce the estimation variance without incurring an extra bias. Otherwise, the gain depends on the nature of $h$ and the candidate models of each learner. 
Specifically, based on the assumptions in the Appendix, we have the following observations. (i)
If $h\in\mathcal{M}_{1}$ is parametric, but at least one of other learners in $\S_t$ does not have a parametric model containing $h$, we have  $\frac{1}{n}-u_{n}^{2}\lesssim\mathcal{R}\le0$.
(ii) If $h\in\cup_{i=2}^{L}\mathcal{M}_{i}\backslash\mathcal{M}_{1}$
is parametric, then $0\le\mathcal{R}\lesssim1-l_{n}^{2}$. (iii) If
$h\notin\cup_{i=1}^{L}\mathcal{M}_{i}$ is parametric or $h$ is non-parametric,
then $l_{n}^{2}-u_{n}^{2}\le\mathcal{R}\lesssim u_{n}^{2}-l_{n}^{2}$.

\paragraph{Data access.}
The SEC algorithm still applies when each learner can only access its own sub-dataset. For example, for learner $l_{i}$, all other learners only need to share their best estimators $\hat{f}_{j}$, $j\neq i$. So each learner $l_{i}$ can calculate and share $\hat{e}_{j\rightarrow i}$ for all $j\neq i$. Then, each learner can access the dissimilarity matrix $v_{ij}$ and apply our SEC algorithm to conduct the clustering. Due to the nature of sharing the dissimilarity matrix, our method requires access to each sub-dataset twice.}

\subsection*{Acknowledgment}
We thank the anonymous reviewers and Editor for their valuable time and comments, which have helped us improve the original manuscript.  
This paper is based upon work supported by the National Science Foundation under grant number ECCS-2038603. The authors report there are no competing interests to declare.

\bibliography{clustering,relatedwork}
\bibliographystyle{agsm}

\end{document}